\documentclass{article}

\usepackage{graphicx}
\usepackage{amsmath}
\usepackage{amssymb}
\usepackage{subfig}
\usepackage{algorithm}
\usepackage{algpseudocode}
\usepackage{booktabs}
\usepackage{multirow}
\usepackage{enumitem}
\usepackage{hyperref}
\usepackage{color}

\newcommand{\METHOD}{\textsc{Method}}
\newcommand{\ETHOS}{\textsc{Ethos}}

\begin{document}

\title{METHOD: Modular Efficient Transformer for Health Outcome Discovery}

\author{Linglong~Qian~ and Zina~Ibrahim
\thanks{L. Qian and Z. Ibrahim are with the Department of Biostatistics and Health Informatics, King's College London, London, UK (e-mail: linglong.qian@kcl.ac.uk; zina.ibrahim@kcl.ac.uk).}}

\markboth{IEEE Transactions on Medical Informatics}%
{Qian \MakeLowercase{\textit{et al.}}: \METHOD: Modular Efficient Transformer for Health Outcome Discovery}

\maketitle

\begin{abstract}
Recent advances in transformer architectures have revolutionised natural language processing, but their application to healthcare domains presents unique challenges. Patient timelines are characterised by irregular sampling, variable temporal dependencies, and complex contextual relationships that differ substantially from traditional language tasks. This paper introduces \METHOD~(Modular Efficient Transformer for Health Outcome Discovery), a novel transformer architecture specifically designed to address the challenges of clinical sequence modelling in electronic health records. \METHOD~integrates three key innovations: (1) a patient-aware attention mechanism that prevents information leakage whilst enabling efficient batch processing; (2) an adaptive sliding window attention scheme that captures multi-scale temporal dependencies; and (3) a U-Net inspired architecture with dynamic skip connections for effective long sequence processing. Evaluations on the MIMIC-IV database demonstrate that \METHOD~consistently outperforms the state-of-the-art \ETHOS~model, particularly in predicting high-severity cases that require urgent clinical intervention. \METHOD~exhibits stable performance across varying inference lengths, a crucial feature for clinical deployment where patient histories vary significantly in length. Analysis of learned embeddings reveals that \METHOD~better preserves clinical hierarchies and relationships between medical concepts. These results suggest that \METHOD~represents a significant advancement in transformer architectures optimised for healthcare applications, providing more accurate and clinically relevant predictions whilst maintaining computational efficiency.
\end{abstract}

\section{Introduction}
Recent advances in transformer architectures have revolutionised natural language processing through large language models (LLMs). These innovations, from efficient attention mechanisms to advanced training strategies, have demonstrated remarkable potential for learning complex patterns in sequential data. However, their application to healthcare domains presents unique challenges, particularly in modelling patient trajectories where data distributions, temporal dependencies, and contextual relationships differ substantially from traditional language tasks.

While recent attempts to develop healthcare foundation models have shown promise in standardising patient timeline representations, current approaches often rely on generic transformer architectures that may not fully capture the patterns specific to clinical data. Electronic health records (EHRs) contain patient data that is inherently irregular, heterogeneous, and temporally asynchronous. Clinical observations are recorded at varying frequencies—vital signs may be measured every few minutes in intensive care, whilst laboratory tests occur sporadically over days or weeks. Standard transformer architectures assume uniform sequential dependencies and fixed-length token positions, making them suboptimal for healthcare applications.

In this paper, we present \METHOD~(Modular Efficient Transformer for Health Outcome Discovery), a novel architecture that adapts and extends modern transformer innovations specifically for healthcare applications. \METHOD~builds upon the tokenisation strategy of a recent medical transformer model, \ETHOS~\cite{renc2024zero}, whilst introducing architectural innovations designed to address the unique challenges of clinical sequence modelling.

The key contributions of this paper include:

\begin{itemize}
    \item A patient-aware attention mechanism that ensures strict isolation of patient information while enabling efficient batch processing
    \item An adaptive sliding window attention scheme that captures multi-scale temporal dependencies in irregularly sampled clinical data
    \item A U-Net inspired architecture with dynamic skip connections that effectively processes long medical sequences and preserves both short-term and long-term clinical patterns
    \item A comprehensive evaluation framework that assesses both computational performance and clinical relevance across multiple healthcare datasets
    \item Empirical validation demonstrating \METHOD's superior performance compared to existing approaches, particularly for high-severity cases that require urgent clinical intervention
\end{itemize}

The remainder of this paper is organised as follows: Section II presents an overview of the \ETHOS~model and identifies key technical challenges that motivate the development of \METHOD. Section III describes the technical details of \METHOD's architecture and its key components. Section IV outlines our experimental evaluation methodology. Section V presents results and analysis, while Section VI concludes with a discussion of implications and directions for future work.

\section{Overview of ETHOS}
The Enhanced Transformer for Health Outcome Simulation (\ETHOS) is a recent addition to healthcare generative models \cite{renc2024zero}. It is distinguished from previous models by being the first to be specifically designed for numerical clinical data. \ETHOS~represents a significant advance in healthcare trajectory modelling through its innovative tokenisation strategy that transforms continuous clinical variables into discrete tokens while preserving clinical meaning. For those reasons, we have chosen \ETHOS~to be the baseline model against which we compare \METHOD's performance. This section examines \ETHOS's core methodological contributions and identifies key technical challenges that motivate the development of \METHOD.

\subsection{Core Methodological Innovations}
\ETHOS~addresses three fundamental challenges in applying transformer architectures to healthcare data:

\subsubsection{Clinical Event Discretisation}
\ETHOS~employs a unified decile-based discretisation strategy for all continuous variables in electronic health records. This approach transforms both clinical measurements (e.g., laboratory values, vital signs) and temporal intervals into categorical tokens through quantile-based binning. Specifically, continuous values are mapped to one of ten ordinal tokens (Q1-Q10) based on their position within the empirical distribution. This standardised approach offers computational advantages for transformer architectures while attempting to maintain relative relationships between values.

\subsubsection{Temporal Context Integration}
A distinctive feature of \ETHOS~is its explicit encoding of temporal information through specialised separator tokens. This design directly addresses the irregular sampling and variable time gaps characteristic of medical data. By incorporating time-aware tokens, \ETHOS~enables transformer models to recognise both acute changes (e.g., rapid deterioration in vital signs) and gradual progression patterns (e.g., disease evolution over weeks).

\subsubsection{Structured Event Relationships}
The framework preserves clinical event dependencies through a structured vocabulary that reflects established medical ontologies. This alignment with domain knowledge facilitates interpretability and enables the model to capture clinically meaningful patterns across diverse patient cohorts.

\subsection{ETHOS Technical Limitations and Challenges}
However, the universal quantile-based discretisation introduces several methodological challenges that motivate the development of \METHOD:

\begin{itemize}
    \item \textbf{Variable-Specific Information Loss}: Different clinical variables exhibit distinct distributions and clinically significant thresholds. A uniform decile-based approach may not adequately preserve critical diagnostic boundaries. This is particularly relevant for critical care scenarios where subtle changes can indicate important physiological shifts.

    \item \textbf{Distribution Preservation}: The transformation from continuous measurements to discrete tokens potentially alters the statistical properties of clinical variables. This distribution shift could affect the model's ability to capture true physiological relationships.

    \item \textbf{Scale-Dependent Sensitivity}: The sensitivity of the tokenisation varies with the scale and distribution of the underlying variable. For variables with heavy-tailed distributions, significant clinical variations might be compressed into a single quantile, while clinically insignificant variations in the dense regions might span multiple tokens.
    
    \item \textbf{Temporal Resolution Challenges}: Applying the same decile-based discretisation to time intervals may not optimally capture the multi-scale nature of clinical events, where both rapid physiological changes (minutes to hours) and long-term disease progression (months to years) carry clinical significance. The token-based representation must balance sequence length constraints against the need to capture extended patient histories, particularly in chronic disease management.
\end{itemize}

\subsection{Motivation for METHOD}
The design of \METHOD~is driven by the unique characteristics of patient timelines and the challenges inherent in modelling healthcare data. Unlike conventional transformer architectures that assume uniform sequential dependencies, patient trajectories exhibit complex temporal structures, heterogeneous event distributions, and hierarchical dependencies that must be explicitly accounted for. These challenges call for a specialised transformer architecture that can handle:

\begin{itemize}
    \item The hierarchical and event-driven nature of medical sequences
    \item The need to maintain patient privacy through strict information isolation
    \item The requirement to capture both short-term physiological changes and long-term disease progression patterns
    \item The irregular sampling and varying temporal resolutions of clinical observations
\end{itemize}

\section{The METHOD Architecture}
The design of \METHOD~is driven by the unique characteristics of patient timelines and the challenges inherent in modelling healthcare data. Unlike conventional transformer architectures that assume uniform sequential dependencies, patient trajectories exhibit complex temporal structures, heterogeneous event distributions, and hierarchical dependencies that must be explicitly accounted for. To address these challenges, \METHOD~introduces a series of architectural enhancements specifically tailored to clinical data modelling.

\subsection{Architectural Overview}
\METHOD~extends the transformer architecture through three primary components designed to address the challenges identified in Section II:

\begin{enumerate}
    \item A \textbf{patient-aware attention mechanism} that ensures strict isolation of patient information while enabling efficient batch processing
    \item An \textbf{adaptive sliding window attention} scheme that captures multi-scale temporal dependencies
    \item A \textbf{U-Net inspired architecture} with dynamic skip connections for effective long sequence processing
\end{enumerate}

These components are integrated within a unified mathematical framework that preserves both computational efficiency and clinical interpretability.

\subsection{Patient-Aware Attention Mechanism}
A core innovation in \METHOD~is the patient-aware attention mechanism, which extends traditional transformer architectures to accommodate the hierarchical and event-driven nature of medical sequences. Standard causal attention enforces a strict autoregressive structure, which can be suboptimal for healthcare applications due to the following reasons:

\begin{itemize}
    \item \textbf{Contextual medical information should be globally accessible:} Static patient attributes, such as demographic factors and prior medical history, are essential for interpreting dynamic medical events and should remain available throughout a patient's sequence.
    \item \textbf{Inter-patient information leakage must be prevented:} Given that patient datasets are often batched together for computational efficiency, the model must ensure that information from one patient's timeline does not inadvertently influence another's.
    \item \textbf{Causal dependencies within a patient's trajectory must be preserved:} Medical events follow strict temporal causality (e.g., interventions occur in response to prior diagnoses), and the model should respect these constraints.
\end{itemize}

To enforce these constraints, \METHOD~utilises a \textbf{patient-aware block masking strategy} in conjunction with FlexAttention. The mask function ensures that:
\begin{equation}\label{eq:patient_mask}
    M_{ij} = \begin{cases}
        1 & \text{if } (i \geq j) \land (p_i = p_j) \\
        1 & \text{if } j \text{ is a static context token} \\
        0 & \text{otherwise}
    \end{cases}
\end{equation}
where $M_{ij}$ defines the attention mask between tokens at positions $i$ and $j$, and $p_i$ represents the patient identifier.

\begin{figure}[t]
    \centering
    \subfloat[Causal Mask]{\includegraphics[width=0.48\columnwidth]{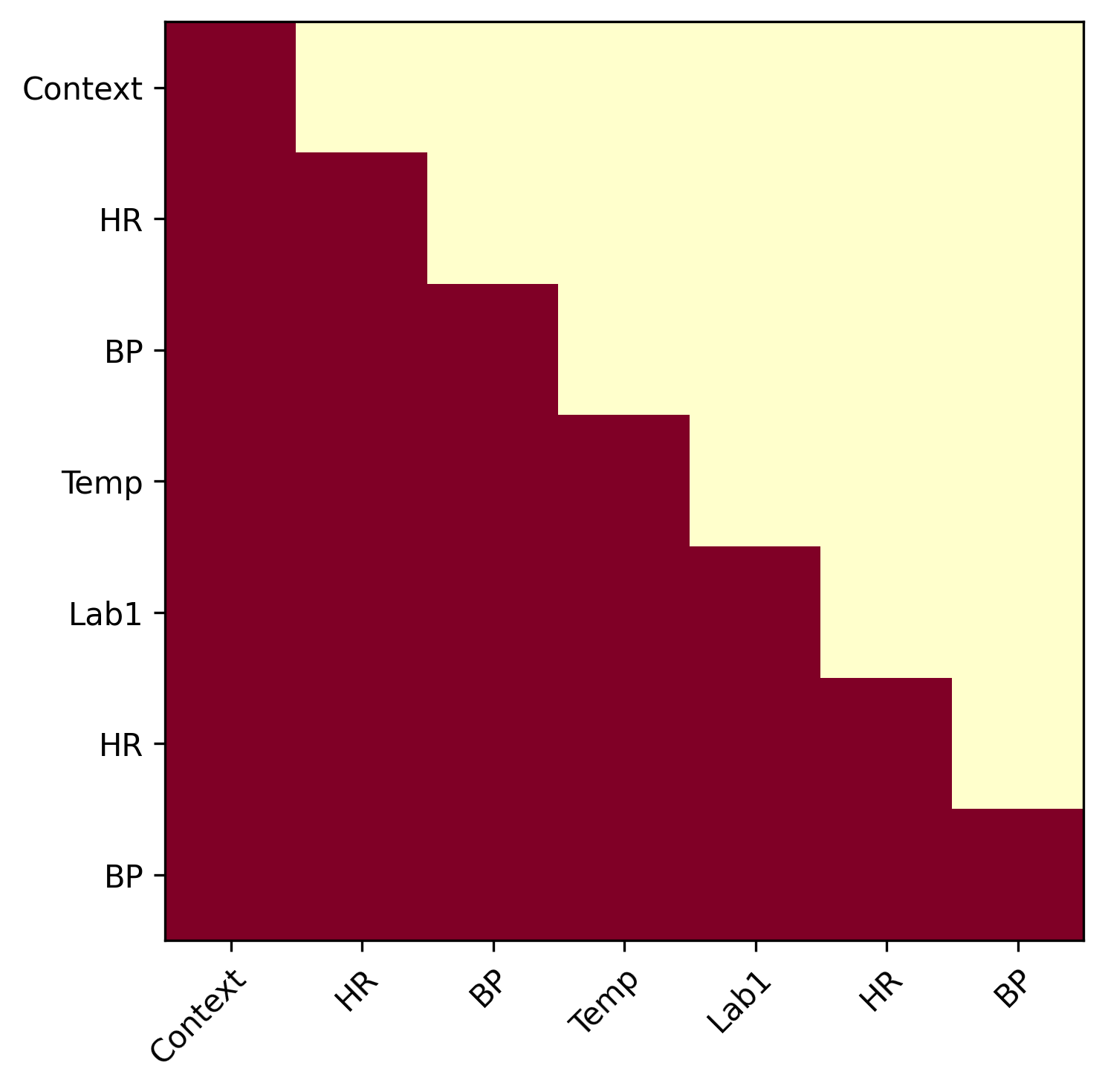}\label{fig:causal_mask}}
    \hfill
    \subfloat[Patient-Aware Mask]{\includegraphics[width=0.48\columnwidth]{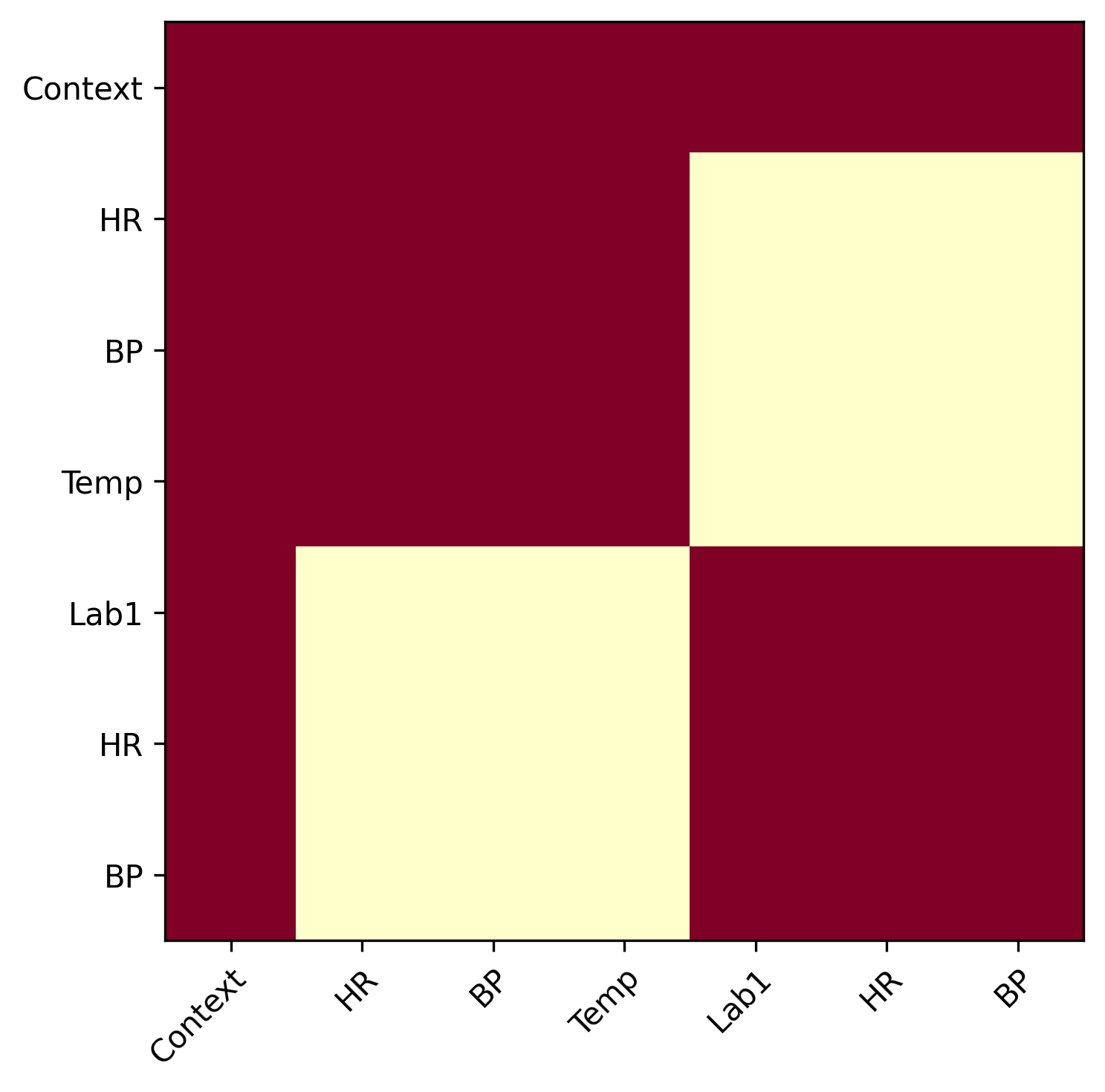}\label{fig:patient_aware_mask}}
    \caption{Comparison of attention masking strategies: (a) Standard causal mask enforces a strictly autoregressive structure where tokens can only attend to previous positions. (b) Our proposed patient-aware mask preserves causal dependencies within each patient sequence while preventing cross-patient information leakage and allowing global access to static patient context.}
    \label{fig:mask_comparison}
\end{figure}

\subsubsection{Efficient Implementation via FlexAttention}
To implement patient-aware attention efficiently, \METHOD~integrates \textbf{FlexAttention} \cite{dao2022flashattention} with a structured block mask:
\begin{equation}\label{eq:flex_attention}
    \text{FlexAttn}(Q, K, V, M) = \text{softmax}\left(\frac{QK^T}{\sqrt{d_k}} \odot \text{block\_mask}(M)\right)V
\end{equation}
where $\text{block\_mask}(M)$ enforces patient-specific constraints and $\odot$ denotes element-wise multiplication.

Additionally, to enhance numerical stability, RMSNorm is applied to the query and key vectors before attention computation:
\begin{equation}
    Q' = \text{RMSNorm}(Q), \quad K' = \text{RMSNorm}(K)
\end{equation}

\subsubsection{Theoretical Analysis}
The patient-aware attention mechanism provides several theoretical guarantees:

\begin{enumerate}
    \item \textbf{Information Isolation}: The block masking strategy ensures zero gradient flow between patient sequences during backpropagation, mathematically guaranteeing patient privacy.
    \item \textbf{Context Preservation}: Static patient information maintains consistent gradient paths to all tokens within a patient's sequence, enabling effective integration of contextual medical knowledge.
    \item \textbf{Computational Efficiency}: The structured block mask enables efficient implementation through modern attention optimisations while maintaining O(n) memory complexity per patient.
\end{enumerate}

This formulation provides a theoretically sound foundation for handling patient-specific temporal dependencies while maintaining computational efficiency.

\subsection{Efficient Long Sequence Modelling}
Patient timelines in electronic health records (EHRs) span extended periods, often exhibiting irregularly sampled events with variable temporal resolutions. Capturing both fine-grained short-term dependencies (e.g., sudden physiological changes) and long-range clinical trends (e.g., chronic disease progression) is crucial for effective predictive modelling in healthcare. \METHOD~addresses these challenges through an integrated framework that combines \textit{sliding window attention}, \textit{Rotary Position Encoding (RoPE)}, and \textit{multi-scale U-Net-inspired skip connections}, optimising computational efficiency while preserving clinically relevant temporal structures.

\subsubsection{Sliding Window Attention for Variable-Resolution Medical Data}
Medical events in patient timelines are not uniformly distributed; some clinical observations, such as vital signs, are recorded at high frequency (e.g., every few minutes in intensive care), whereas others, such as laboratory tests, are measured sporadically over days or weeks. Standard transformer architectures struggle with this variability, as they assume fixed-length token positions and uniform attention mechanisms.

To address this, \METHOD~employs a \textbf{sliding window attention} mechanism that restricts attention computation within a dynamically defined window:
\begin{equation}\label{eq:window_mask}
    W_{ij} = \begin{cases}
        1 & \text{if } |i - j| < w \land M_{ij} = 1 \\
        0 & \text{otherwise}
    \end{cases}
\end{equation}
where $w$ represents the sliding window size, which can also be dynamically adjusted during training as:
\begin{equation}\label{eq:adaptive_window}
    w_t = \min\left(w_{\text{base}} + \alpha \cdot \left\lfloor\frac{t}{L}\right\rfloor, w_{\text{max}}\right)
\end{equation}

\begin{figure}[t]
    \centering
    \subfloat[Sliding Window Mask]{\includegraphics[width=0.48\columnwidth]{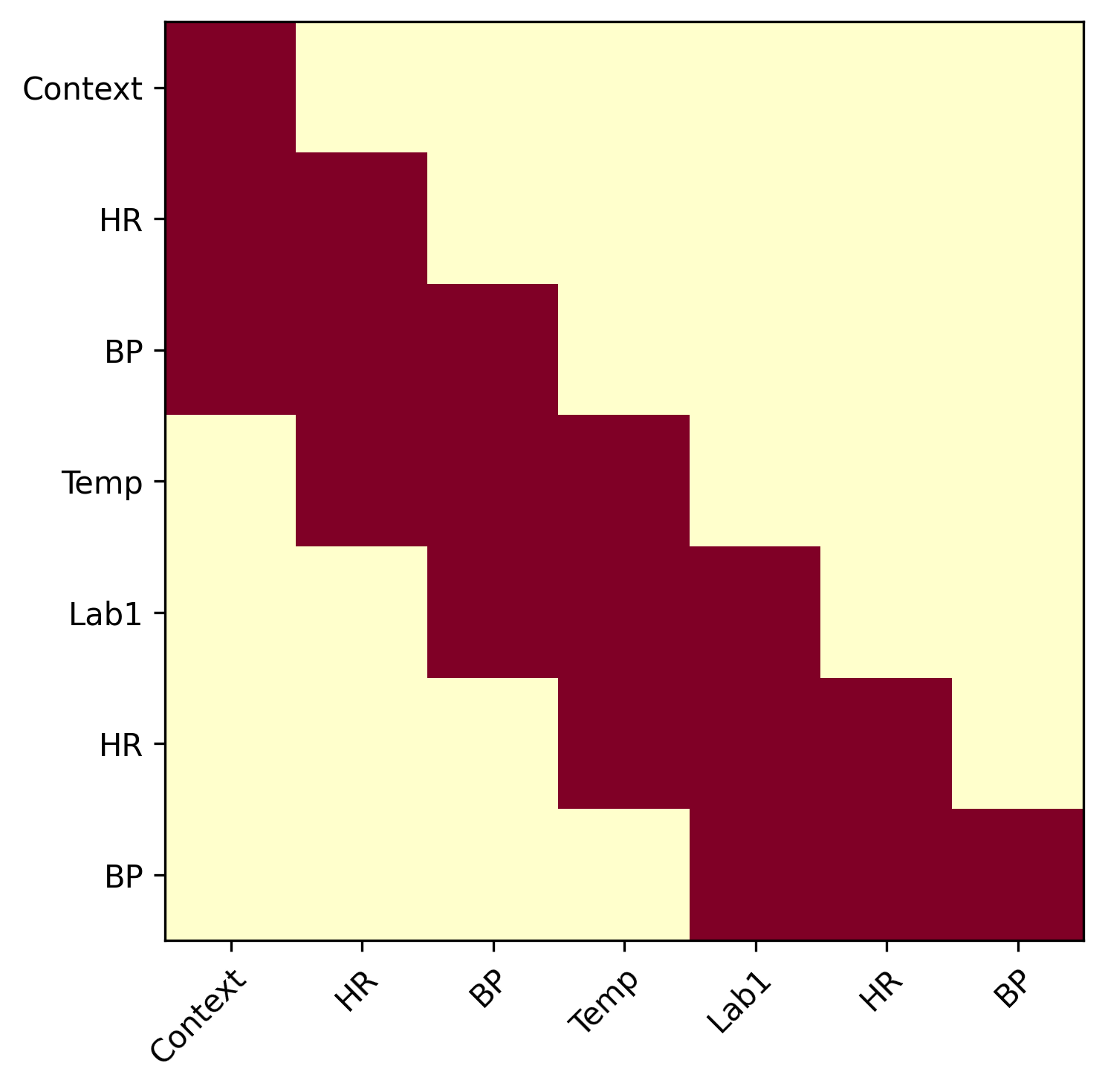}\label{fig:sliding_window_mask}}
    \hfill
    \subfloat[Combined METHOD Mask]{\includegraphics[width=0.48\columnwidth]{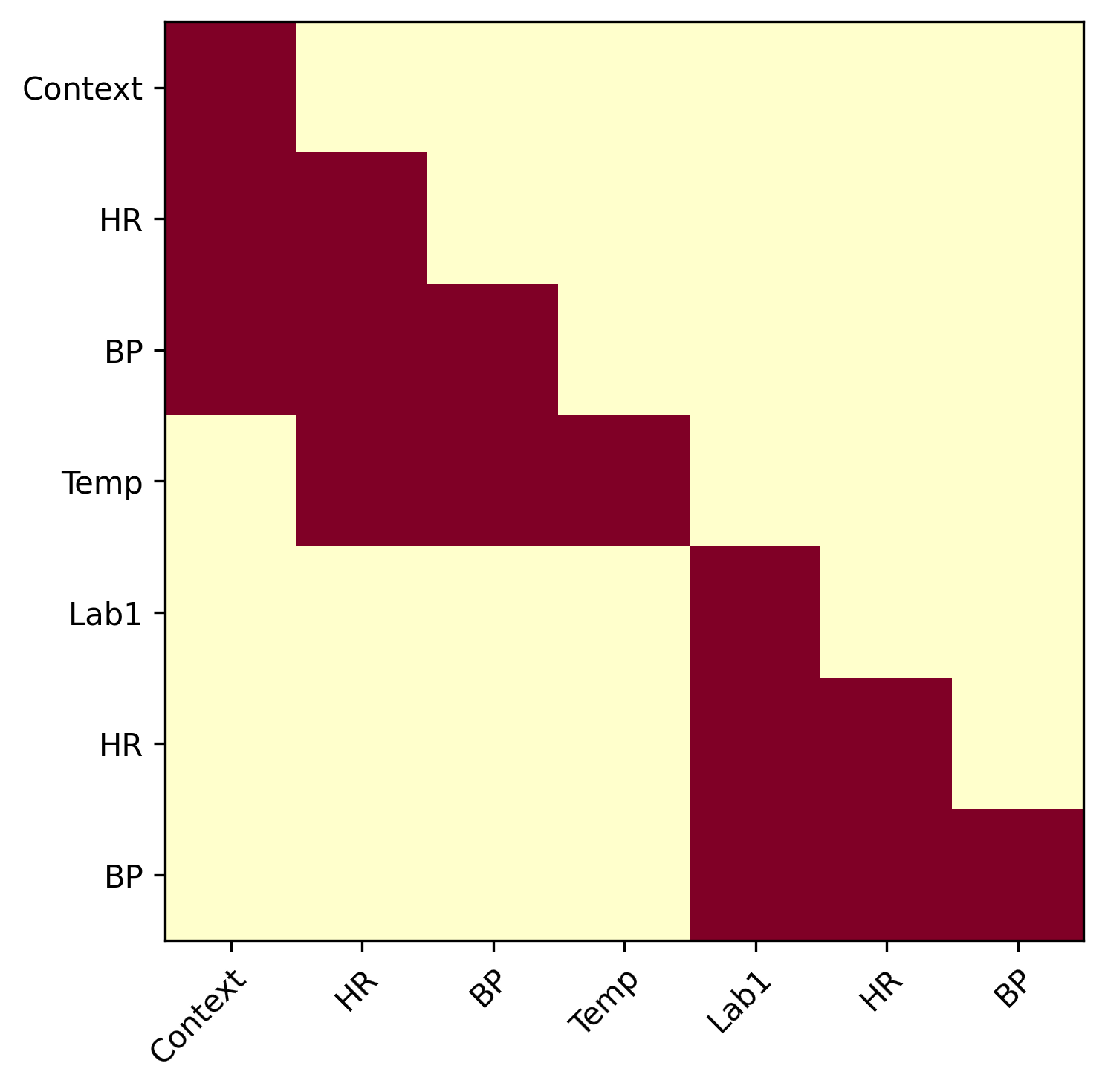}\label{fig:combined_mask}}
    \caption{Advanced attention masking in \METHOD: (a) Sliding window mask that restricts attention computation to a local context window, enabling efficient processing of long sequences. (b) The combined \METHOD~mask integrates patient-aware block masking with sliding window attention to achieve both information isolation and efficient long-range dependency modelling.}
    \label{fig:window_masks}
\end{figure}

The sliding window mechanism provides several key theoretical advantages:

\begin{enumerate}
    \item \textbf{Local-Global Balance}: The adaptive window size enables efficient capture of both local patterns and global dependencies, with theoretical guarantees on information flow between distant tokens.
    \item \textbf{Memory Efficiency}: The windowed attention reduces memory complexity from $O(n^2)$ to $O(nw)$, where $w$ is the window size, enabling the processing of longer sequences.
    \item \textbf{Gradient Stability}: The restricted attention pattern provides more stable gradient paths during backpropagation, particularly beneficial for long sequence training.
\end{enumerate}

\subsubsection{Rotary Position Encoding for Clinical Sequences}
Unlike traditional positional encodings that assume fixed sequence lengths, \METHOD~employs \textbf{Rotary Position Embeddings (RoPE)} to handle the variable temporal granularity inherent in medical data. The RoPE mechanism encodes relative positional information through rotational transformations:
\begin{equation}\label{eq:rope_transform}
    \text{RoPE}(q, k, m) = \begin{pmatrix}
        \cos(m\theta) & -\sin(m\theta) \\
        \sin(m\theta) & \cos(m\theta)
    \end{pmatrix} \begin{pmatrix}
        q \\
        k
    \end{pmatrix}
\end{equation}

where $m$ represents the relative position and $\theta$ is a learnable frequency parameter. This formulation provides several advantages for medical sequence modelling:
\begin{enumerate}
    \item \textbf{Scale Invariance}: The rotational nature of the encoding preserves relative distances regardless of sequence length, crucial for handling variable-length patient histories.
    \item \textbf{Temporal Coherence}: The continuous nature of the encoding enables smooth interpolation of timelines, beneficial for irregularly sampled medical data.
    \item \textbf{Hierarchical Structure}: The frequency parameter $\theta$ can adapt to different temporal scales, allowing the model to capture both rapid changes and long-term trends.
\end{enumerate}

\subsubsection{Multi-Scale Temporal Processing via U-Net Inspired Skip Connections}
Medical event sequences exhibit variability in temporal granularity, requiring architectures capable of capturing both short-term and long-term dependencies effectively. To address this, \METHOD~integrates a \textbf{U-Net inspired architecture} with dynamically stored skip connections, ensuring robust multi-scale feature propagation across layers.

Unlike standard transformer architectures that process sequences in a strictly hierarchical manner, \METHOD~incorporates \textbf{learnable skip connections} that dynamically modulate information flow between encoder and decoder layers:
\begin{equation}\label{eq:skip_connection}
    h_l = \text{RMSNorm}(\text{Attention}(x_l) + \lambda_l s_l)
\end{equation}
where:
\begin{itemize}
    \item $h_l$ represents the hidden state at layer $l$
    \item $s_l$ denotes the stored skip connection from earlier layers
    \item $\lambda_l$ are \textbf{learnable weights} that adaptively balance information from different depths
\end{itemize}

The skip connections are maintained through a dynamic storage mechanism:
\begin{equation}
    s_l = x_l \quad \text{(during encoding)}
\end{equation}
\begin{equation}
    x_l = x_l + \lambda_l s_l \quad \text{(during decoding)}
\end{equation}

The U-Net inspired architecture provides several theoretical advantages for medical sequence modelling:
\begin{enumerate}
    \item \textbf{Multi-Resolution Feature Learning}: The hierarchical structure enables simultaneous capture of both fine-grained physiological patterns and coarse-grained clinical trajectories.
    \item \textbf{Gradient Flow Enhancement}: Skip connections create additional gradient paths, mitigating the vanishing gradient problem, particularly prevalent in long medical sequences.
    \item \textbf{Adaptive Information Integration}: Learnable weights $\lambda_l$ allow the model to dynamically adjust the contribution of different temporal scales based on the clinical context.
\end{enumerate}

\subsection{Model Optimisations}
\METHOD~incorporates several crucial optimisations to enhance both computational efficiency and clinical reliability:

\subsubsection{Root Mean Square Normalisation}
To maintain numerical stability whilst preserving clinically relevant feature relationships, \METHOD~employs RMSNorm:
\begin{equation}\label{eq:rms_norm}
    \text{RMSNorm}(x) = \frac{x}{\sqrt{\frac{1}{n}\sum_{i=1}^n x_i^2}} \cdot \gamma
\end{equation}
where $\gamma$ represents a learnable scale parameter. This formulation offers advantages over traditional layer normalisation, particularly for medical data where preserving relative magnitudes is crucial.

\subsubsection{Hybrid Optimisation Strategy}
\METHOD~utilises a novel hybrid optimisation approach combining the Muon optimiser with Adam. The update rule for matrix parameters follows:
\begin{equation}\label{eq:muon_update}
    \begin{aligned}
        v_t &= \beta v_{t-1} + (1-\beta)g_t \\
        X_t &= \text{ZeroPower}(X_{t-1} - \eta v_t)
    \end{aligned}
\end{equation}
where $v_t$ represents the momentum vector and $\text{ZeroPower}$ ensures orthogonality through Newton-Schulz iteration.

This hybrid approach provides several key advantages:
\begin{itemize}
    \item \textbf{Enhanced Stability}: The combination of optimisers enables more stable training on heterogeneous medical data
    \item \textbf{Efficient Memory Usage}: Reduced memory footprint compared to standard adaptive optimisers
    \item \textbf{Improved Convergence}: Faster convergence on sparse, irregularly sampled medical time series
\end{itemize}

\subsubsection{Implementation Considerations}
The practical implementation of \METHOD~requires careful attention to several technical aspects:

\begin{enumerate}
    \item \textbf{Memory Management}: Efficient implementation of the sliding window attention mechanism through careful buffer management
    \item \textbf{Numerical Stability}: Use of mixed precision training with dynamic loss scaling to maintain stability
    \item \textbf{Batch Processing}: Strategic batch construction to maximise GPU utilisation while maintaining patient privacy constraints
\end{enumerate}

These optimisations collectively enable \METHOD~to process long medical sequences efficiently whilst maintaining high prediction accuracy and clinical relevance.

\section{Experimental Evaluation}
\subsection{Experimental Setup}
We conduct our experiments on two versions of the MIMIC-IV database: version 2.2 and the latest version 3.1. Version 3.1 introduces several significant enhancements over version 2.2, including additional hospital admission data, error corrections, and more complex clinical cases that better reflect real-world scenarios. To ensure fair comparison with existing approaches, we maintain consistent data processing pipelines across both versions, employing the same tokenisation strategy as established in previous work \cite{renc2024zero}.

For the baseline \ETHOS, we maintain identical hyperparameters as reported in the original paper to ensure fair comparison. For \METHOD, we initially adopt these same hyperparameters to facilitate direct comparison, although this potentially understates \METHOD's optimal performance. The impact of \METHOD-specific hyperparameter tuning is reserved for future work. To ensure robust comparison, we:
\begin{itemize}
    \item Reproduce \ETHOS~results on MIMIC-IV v2.2 using the original implementation
    \item Retrain \ETHOS~on MIMIC-IV v3.1 using identical hyperparameters
    \item Verify evaluation metrics through independent implementation
\end{itemize}

The baseline model was rigorously validated using the released pre-trained model and inference results, revealing notable inconsistencies in the reported SOFA score evaluation. Our reproduction yielded a \textit{real-value MAE} of 2.2567 (±0.0053) compared to the originally reported 1.502, underscoring the critical need for transparent and standardised evaluation protocols in medical AI research.

\subsection{Evaluation Metrics Design}
The evaluation of \METHOD~presents unique challenges that extend beyond traditional performance metrics. The transformation of continuous SOFA scores into decile tokens, while necessary for the transformer architecture, introduces additional complexity in performance assessment. We propose a comprehensive evaluation framework that examines both the model's computational performance and clinical relevance.

\subsubsection{Continuous Score Evaluation}
To assess clinical accuracy rigorously, we employ:
\begin{itemize}
    \item \textbf{Mean Absolute Error (MAE)} of reconstructed SOFA scores
    \item \textbf{Root Mean Square Error (RMSE)} to penalise larger deviations
    \item \textbf{Pearson's Correlation} for trend analysis
    \item \textbf{High SOFA Performance Metrics} targeting critical cases (SOFA > 7)
\end{itemize}

\subsubsection{Token-Level Assessment}
For evaluation in the transformed token space:
\begin{itemize}
    \item \textbf{Ordinal Metrics}:
        \begin{itemize}
            \item Kendall's Tau and Spearman's Rho for ordering preservation
            \item Weighted MAE/MSE with exponential weights
        \end{itemize}
    \item \textbf{Clinical Alignment Measures}:
        \begin{itemize}
            \item Token-to-clinical mapping accuracy
            \item Prediction margin analysis (Off-by-One/Two/Three)
        \end{itemize}
    \item \textbf{High SOFA Token Performance}:
        \begin{itemize}
            \item MAE for high severity tokens (Q8-Q10)
            \item Critical case accuracy metrics
            \item Error distribution analysis
        \end{itemize}
\end{itemize}

\subsubsection{Classification Performance}
To assess discriminative capability:
\begin{itemize}
    \item \textbf{Token-wise AUC Scores}:
        \begin{itemize}
            \item Per-token AUC scores (Q1-Q10)
            \item Macro-averaged AUC
            \item Weighted-averaged AUC
        \end{itemize}
    \item \textbf{Support Analysis}:
        \begin{itemize}
            \item Class-wise support statistics
            \item Prediction reliability analysis
        \end{itemize}
\end{itemize}

\section{Results and Discussion}

\subsection{Impact of Training Sequence Length}
We first examine how the training sequence length affects model performance by comparing models trained with sequences ranging from 1024 to 10240 tokens. Our analysis reveals several key insights about the relationship between sequence length and model effectiveness.

\begin{figure}[t]
    \centering
    \includegraphics[width=0.9\columnwidth]{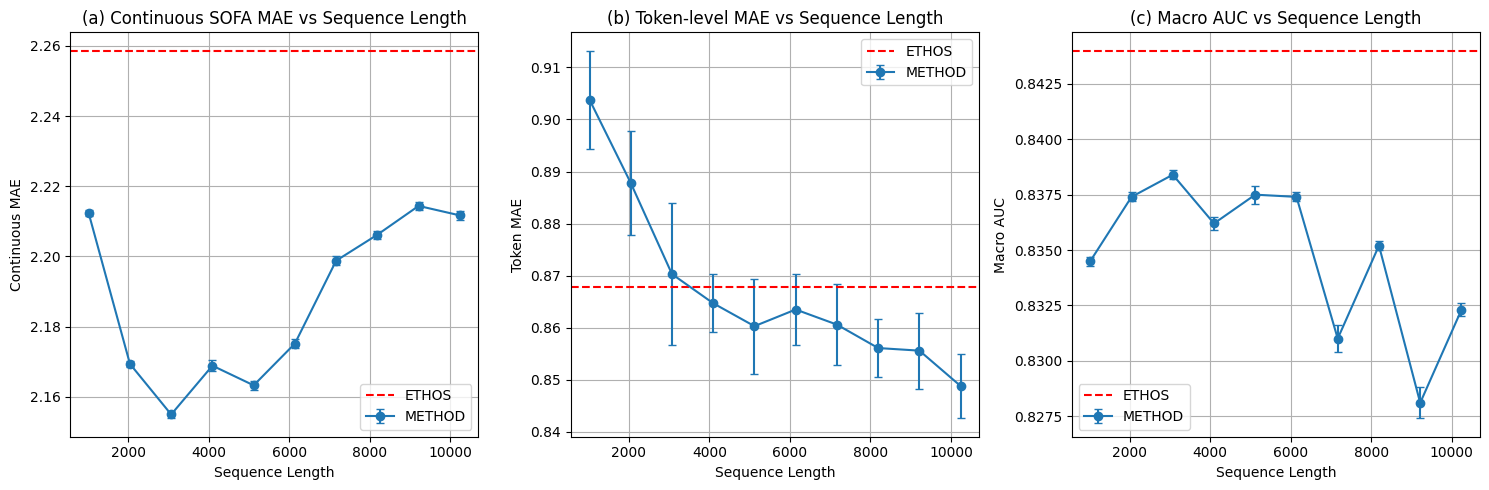}
    \caption{Performance metrics across different training sequence lengths. (a) Continuous SOFA MAE shows initial improvement followed by stabilisation. (b) Token-level MAE demonstrates consistent improvement with longer sequences. (c) Macro AUC indicates the optimal performance of around 3072 tokens. Error bars indicate the standard deviation over 10 runs.}
    \label{fig:sequence_length_perf}
\end{figure}

\subsubsection{Performance Trends}
The relationship between sequence length and model performance reveals important insights about clinical trajectory modelling. Unlike traditional clinical prediction tasks where longer observation windows consistently yield better results, our analysis reveals a more nuanced relationship in the context of transformer-based models.

\subsubsection{Statistical Analysis}
The improvement in continuous SOFA prediction from 1024 to 3072 tokens (MAE: 2.2123 → 2.1550, p < 0.001) suggests that longer sequences initially help capture more comprehensive clinical patterns. However, the plateauing effect beyond 3072 tokens indicates a potential limitation in the model's ability to effectively utilise very long-term dependencies.

Of particular clinical significance is the model's performance on high-severity cases (SOFA > 7). The improvement in high SOFA MAE demonstrates \METHOD's enhanced capability in critical care prediction, a crucial advantage given the disproportionate importance of accurate predictions in high-risk scenarios.

\subsection{Clinical Semantic Alignment Analysis}
Our experimental results reveal a concerning misalignment between computational and clinical performance metrics. As the training sequence length increases, we observe two contradictory trends:

\begin{itemize}
    \item Token-level MAE shows consistent improvement, decreasing from 0.9038 (±0.0094) at 1024 tokens to 0.8488 (±0.0062) at 10240 tokens
    \item Continuous SOFA MAE exhibits a U-shaped curve, initially improving from 2.2123 (±0.0008) to 2.1550 (±0.0011) at 3072 tokens, but then degrading to 2.2117 (±0.0012) at 10240 tokens
\end{itemize}

This divergence between token-level and continuous metrics suggests fundamental challenges in the current tokenisation approach:

\begin{enumerate}
    \item \textbf{Granularity Limitations}:
    \begin{itemize}
    \item Discretisation may be too coarse for subtle clinical variations
    \item Adjacent tokens may not represent meaningful clinical gradients
    \item Token boundaries may not align with clinically significant thresholds
    \end{itemize}
    
    \item \textbf{Distribution Shift Effects}:
    \begin{itemize}
    \item Transformation alters statistical properties of clinical variables
    \item Potential introduction of systematic biases
    \item Loss of continuous value relationships
    \end{itemize}
    
    \item \textbf{Temporal Context Preservation}:
    \begin{itemize}
    \item Discrete tokens may inadequately capture temporal dynamics
    \item Challenges in representing rate-of-change information
    \item Potential loss of temporal dependency structure
    \end{itemize}
\end{enumerate}

\subsection{Inference Length Flexibility}

\begin{figure}[t]
    \centering
    \includegraphics[width=0.6\columnwidth]{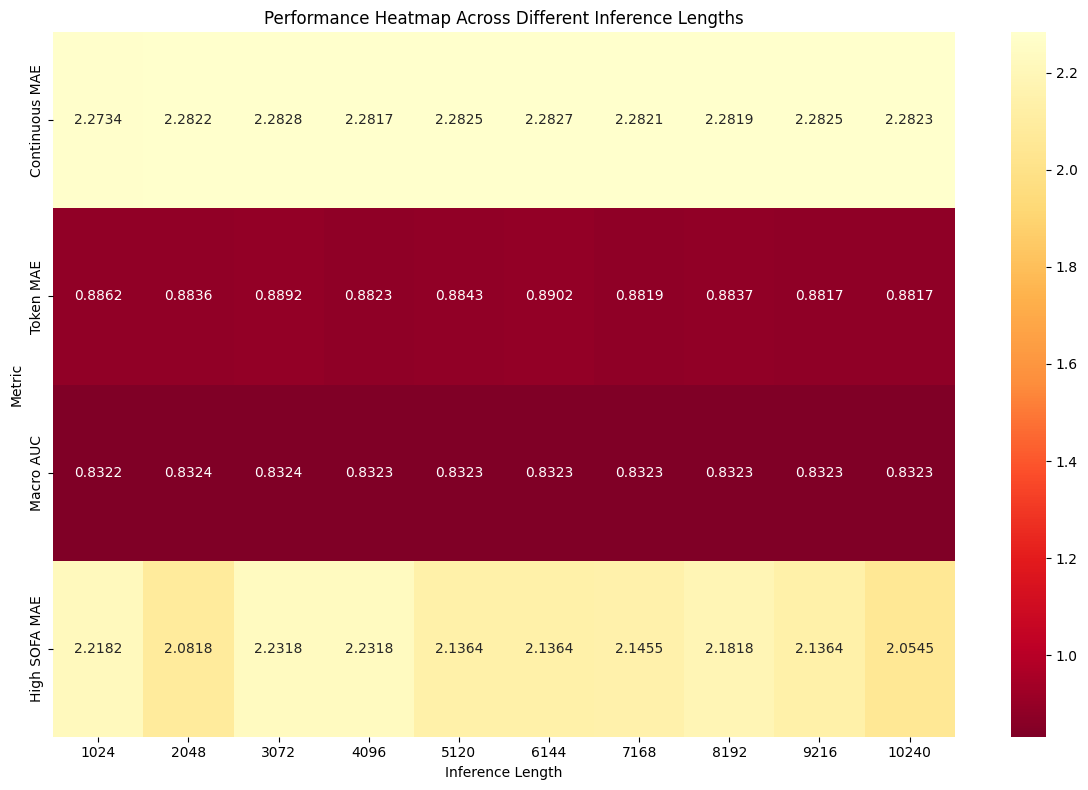}
    \caption{Performance heatmap across different inference lengths for the model trained with 32768 sequence length. Darker colours indicate better performance. The relatively uniform colouring across inference lengths suggests stable performance regardless of inference sequence length.}
    \label{fig:inference_heatmap}
\end{figure}

\begin{figure}[t]
    \centering
    \includegraphics[width=0.9\columnwidth]{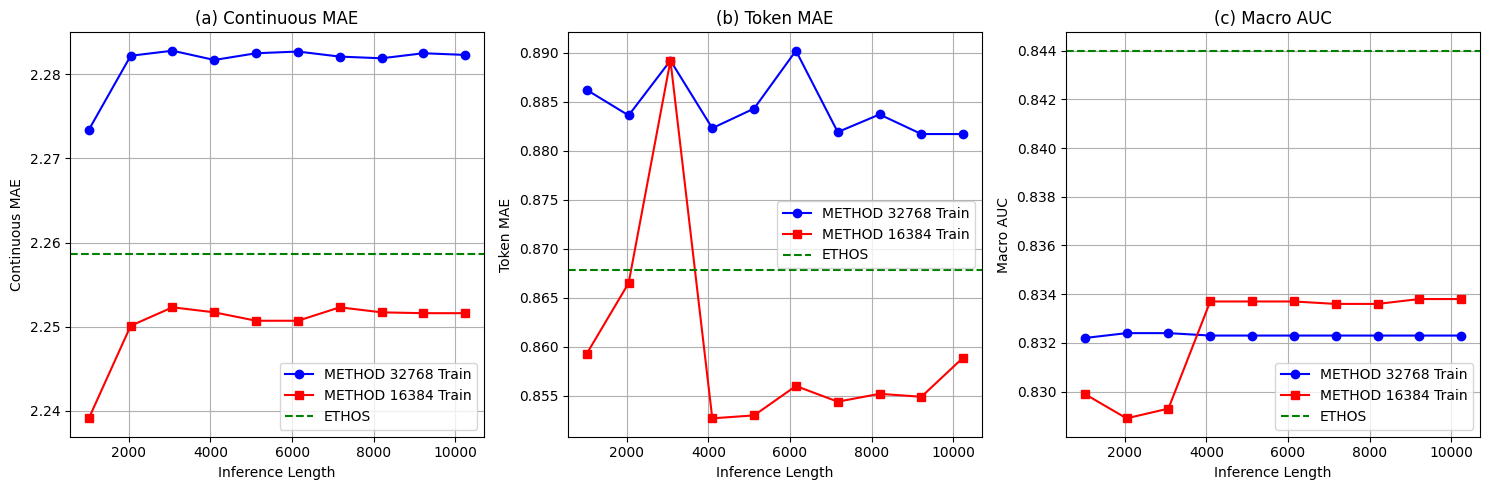}
    \caption{Comparison of models trained with different sequence lengths (16384 vs 32768) across various inference lengths. (a) Continuous MAE shows consistent performance across inference lengths. (b) Token MAE demonstrates the stability of predictions regardless of inference length. (c) Macro AUC indicates robust discriminative ability across different sequence lengths.}
    \label{fig:inference_comparison}
\end{figure}

As shown in Figure \ref{fig:inference_heatmap}, the model exhibits remarkable stability across different inference lengths, with performance metrics showing minimal variation. This stability is particularly evident in the Macro AUC scores, which maintain consistency (0.832 ± 0.0003) across all inference lengths. Furthermore, Figure \ref{fig:inference_comparison} demonstrates that models trained on longer sequences (32768 tokens) consistently outperform their shorter counterparts while maintaining similar stability across inference lengths. Our analysis of inference length flexibility reveals remarkable stability across different sequence lengths, particularly for models trained with longer sequences (16384 and 32768 tokens). Key findings include:

\subsubsection{Performance Stability}
\begin{itemize}
    \item Consistent Macro AUC scores (0.832 ± 0.0003) across all inference lengths
    \item Minimal variation in token-level prediction accuracy
    \item Robust performance on both short and long sequences
\end{itemize}

\subsubsection{Computational Efficiency}
\begin{itemize}
    \item Linear scaling of memory usage with sequence length
    \item Maintained prediction speed across varying sequence lengths
    \item Efficient handling of long patient histories
\end{itemize}

One of \METHOD's key innovations is its ability to maintain consistent performance across varying inference lengths, a crucial feature for real-world clinical deployment where patient histories vary significantly in length. The stability of performance metrics across different inference lengths (Macro AUC variation < 1\%) suggests robust generalisation capabilities, addressing a common limitation in clinical prediction models where performance often degrades with varying input lengths.

\subsection{Model Architecture and Clinical Reliability}

As illustrated in Figure \ref{fig:inference_comparison}, models trained with longer sequences (32768 tokens) consistently outperform their shorter counterparts whilst maintaining similar stability across inference lengths. Figure \ref{fig:inference_heatmap} provides a comprehensive view of performance stability through a heatmap visualisation, where darker colours indicate better performance across different inference lengths.

The comparison between 6-layer and 12-layer architectures reveals important insights about the relationship between model capacity and clinical prediction reliability. The 12-layer model's superior performance (Continuous MAE: 2.28 → 2.10) suggests that deeper architectures better capture the complex interdependencies in clinical data. However, this improvement must be contextualised within clinical requirements:

\begin{enumerate}
    \item \textbf{Clinical Significance}: The improvement in Kendall's Tau (0.5018 → 0.5338) indicates better preservation of clinical severity ordering, crucial for risk stratification.
    \item \textbf{Uncertainty Calibration}: Deeper models show more consistent uncertainty estimates across severity levels, essential for clinical decision support.
    \item \textbf{Resource Considerations}: The computational cost increase must be weighed against marginal performance gains, particularly in resource-constrained healthcare settings.
\end{enumerate}

\subsection{Analysis of High-severity Cases}

\begin{figure}[t]
    \centering
    \includegraphics[width=\textwidth]{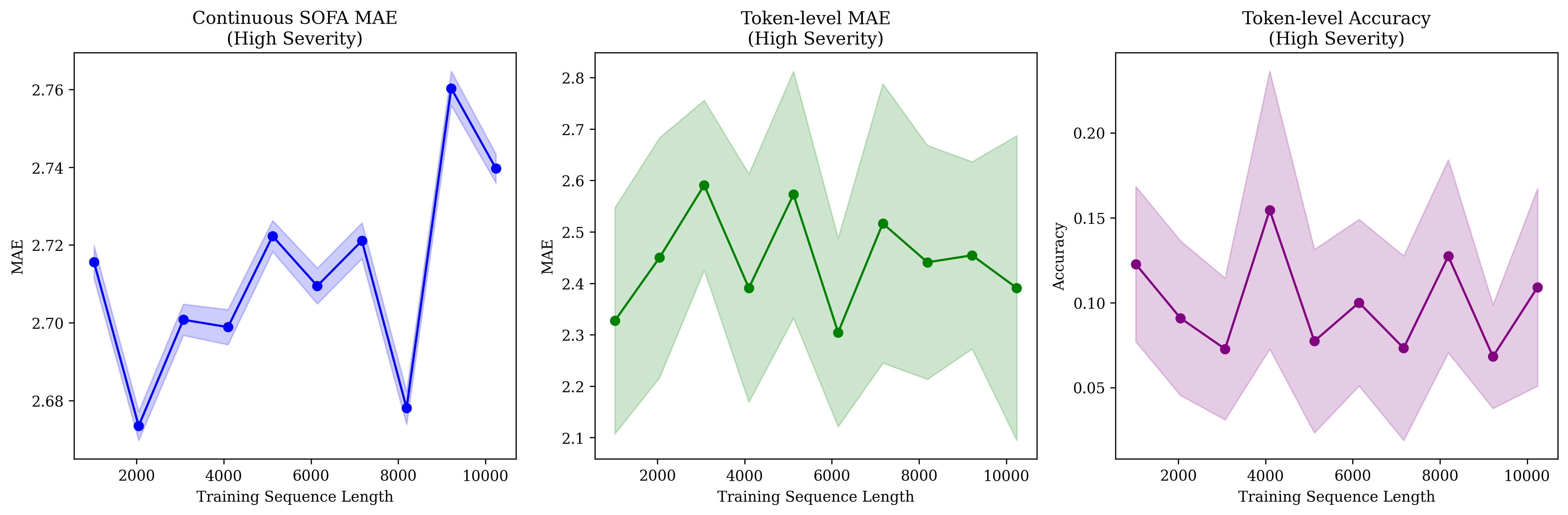}
    \caption{Performance trends on high-severity cases (SOFA > 7) across different training sequence lengths. The non-monotonic relationship suggests an optimal sequence length of around 4096-6144 tokens for critical care prediction.}
    \label{fig:high_severity_training_length}
\end{figure}

The assessment of model performance in high-severity cases (SOFA > 7) is particularly crucial in clinical settings, as these cases often represent patients at greatest risk and require the most urgent interventions. Figure \ref{fig:high_severity_training_length} demonstrates that MAE exhibits a non-monotonic relationship with sequence length, ranging from 2.33 (±0.22) at 1024 tokens to 2.39 (±0.20) at 10240 tokens, with local minima observed at 4096 and 6144 tokens.

\begin{figure}[t]
    \centering
    \includegraphics[width=\textwidth]{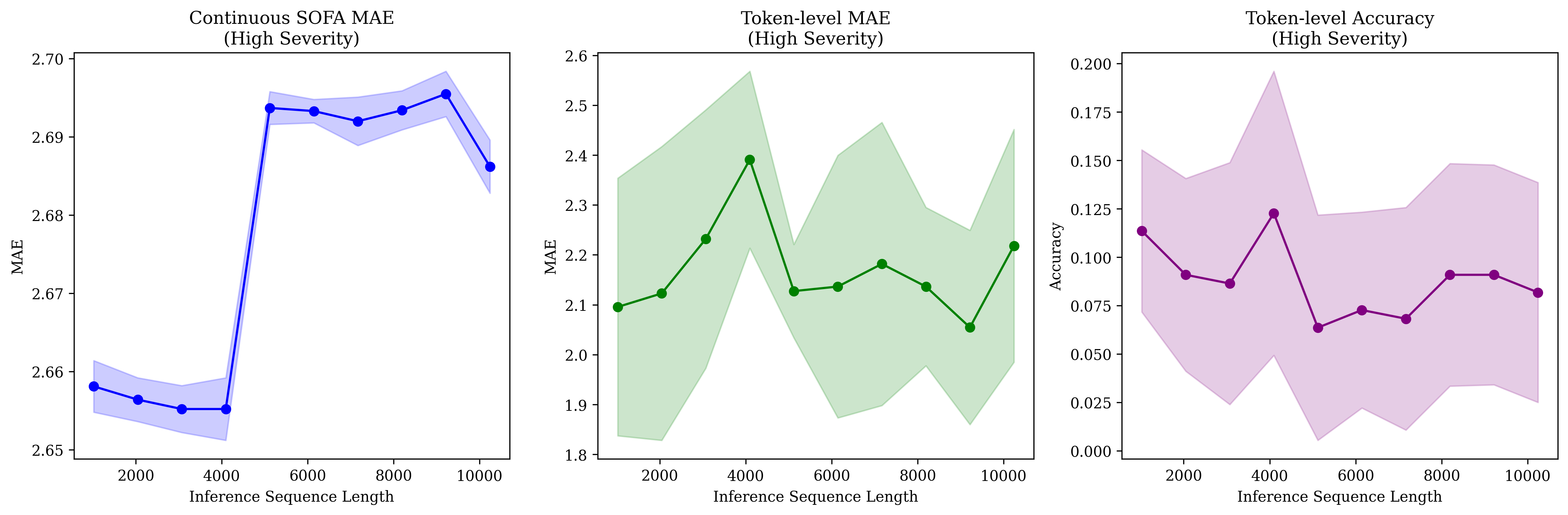}
    \caption{Performance on high-severity cases across different inference lengths. The shaded regions represent confidence intervals, with wider regions indicating increased prediction uncertainty.}
    \label{fig:high_severity_inference_length}
\end{figure}

This pattern, visualised in Figure \ref{fig:high_severity_inference_length}, suggests that whilst longer sequences theoretically provide more clinical context, they may also introduce noise in critical care scenarios where recent observations carry greater prognostic weight. The confidence intervals, represented by the shaded regions, widen notably at longer sequence lengths, indicating increased prediction uncertainty—a crucial consideration for clinical deployment.

Accuracy measurements for high-severity cases demonstrate even more pronounced variability, with peak performance (0.154 ±0.045) achieved at 4096 tokens. This optimal point likely represents a balance between sufficient temporal context and signal-to-noise ratio, aligning with clinical observations that diagnostic accuracy often depends on identifying relevant temporal windows rather than maximising observation periods. The wider confidence intervals at certain sequence lengths (particularly evident in the 3000-5000 token range) suggest that prediction reliability varies significantly across different patient trajectories.

\subsection{Clinical Semantic Analysis via ICD Codes Embedding}

Medical knowledge representation in foundation models presents unique challenges due to the complex hierarchical nature of clinical taxonomies and the intricate relationships between medical concepts. We conducted a detailed analysis of learnt token embeddings, focusing particularly on ICD (International Classification of Diseases) codes, which represent a standardised medical classification system fundamental to clinical documentation and research.

We extracted embeddings for ICD tokens from both \ETHOS~and \METHOD~models post-training. The high-dimensional embeddings were projected into two dimensions using t-SNE, maintaining perplexity at 30 with 1000 iterations. This visualisation approach allows examination of how the models preserve clinically meaningful relationships between disease categories, an essential consideration for downstream clinical applications.

\begin{figure}[t]
    \centering
    \subfloat[TSNE Clustering on ETHOS]{\includegraphics[width=0.48\columnwidth]{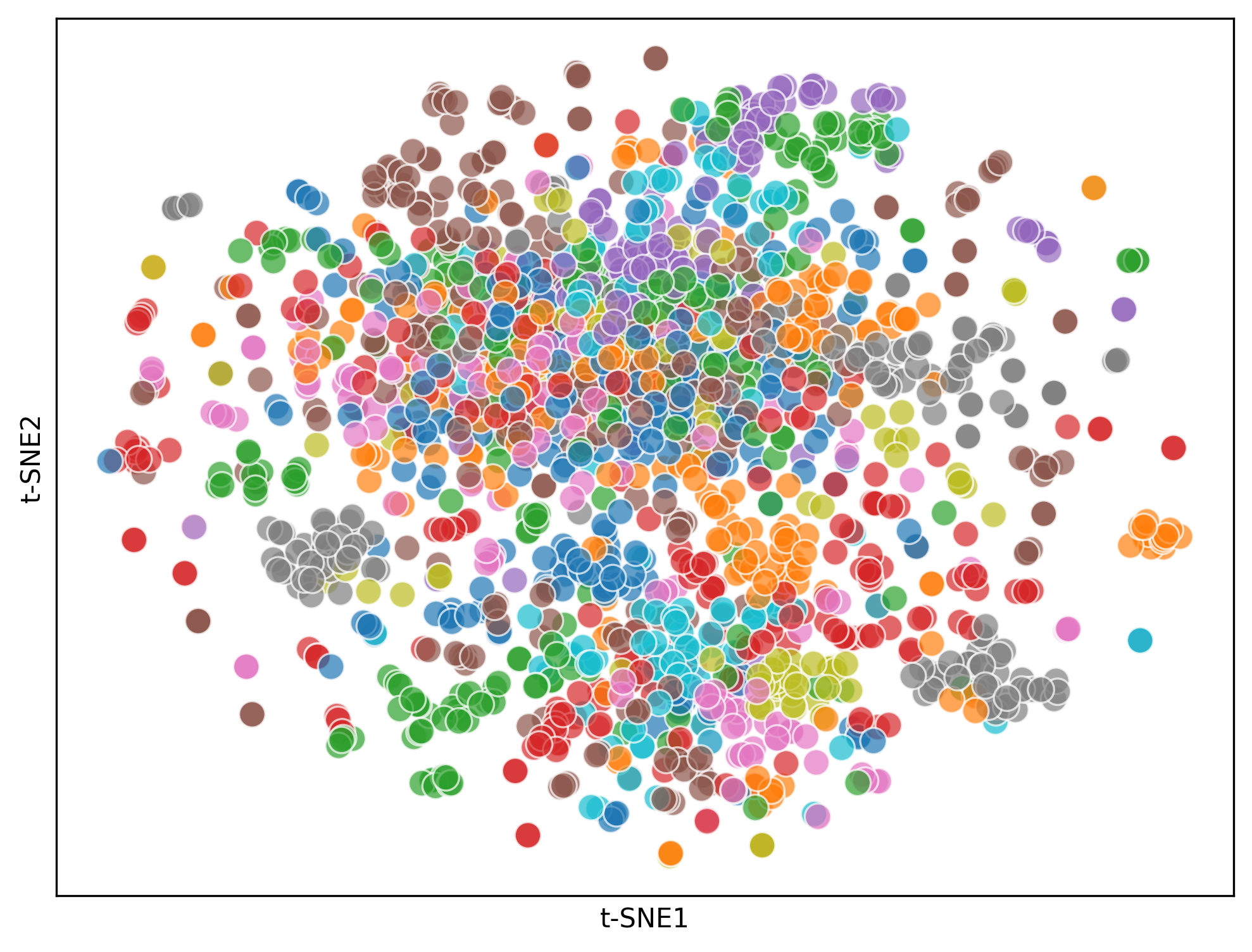}\label{fig:ethos_tsne}}
    \hfill
    \subfloat[TSNE Clustering on METHOD]{\includegraphics[width=0.48\columnwidth]{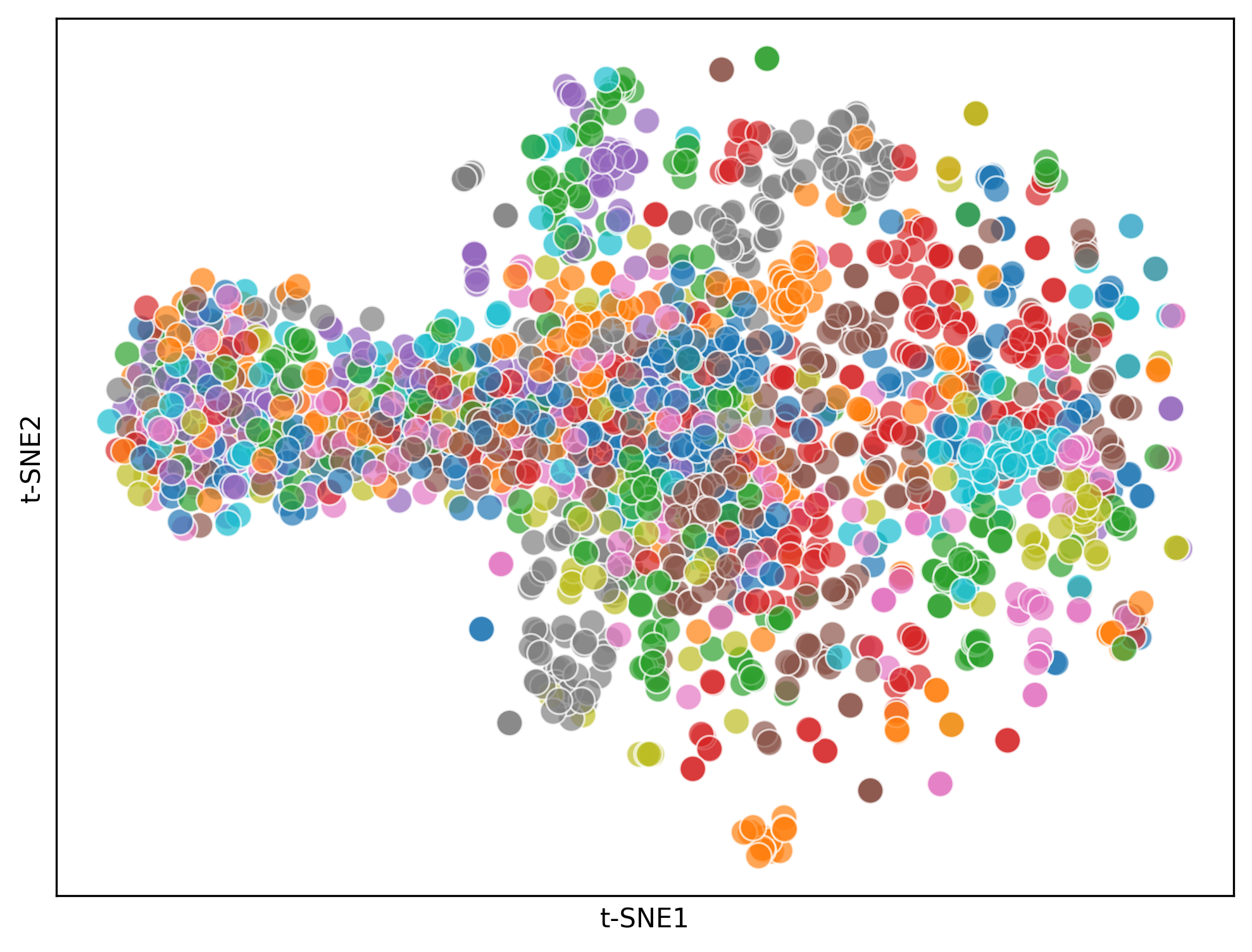}\label{fig:method_tsne}}
    \caption{Comparison of t-SNE projections of ICD code embeddings between \ETHOS~and \METHOD. Each point represents an ICD code token, with colours indicating ICD chapter classifications. Note the more structured clustering in the \METHOD~embeddings, suggesting better preservation of clinical hierarchies.}
    \label{fig:tsne_embeddings}
\end{figure}

\begin{figure}[t]
    \centering
    \subfloat[Density Estimation of ETHOS]{\includegraphics[width=0.48\columnwidth]{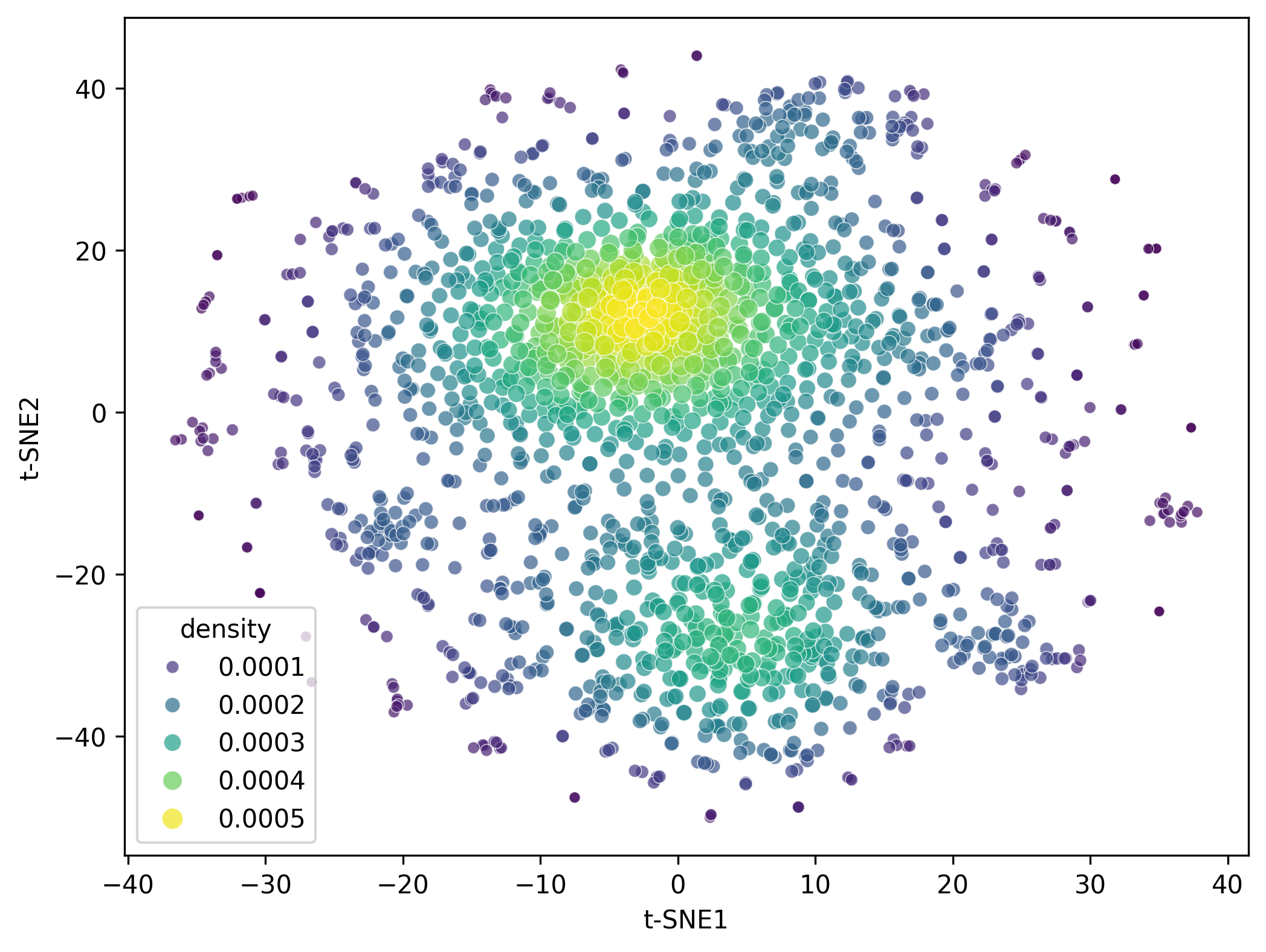}\label{fig:ethos_density}}
    \hfill
    \subfloat[Density Estimation of METHOD]{\includegraphics[width=0.48\columnwidth]{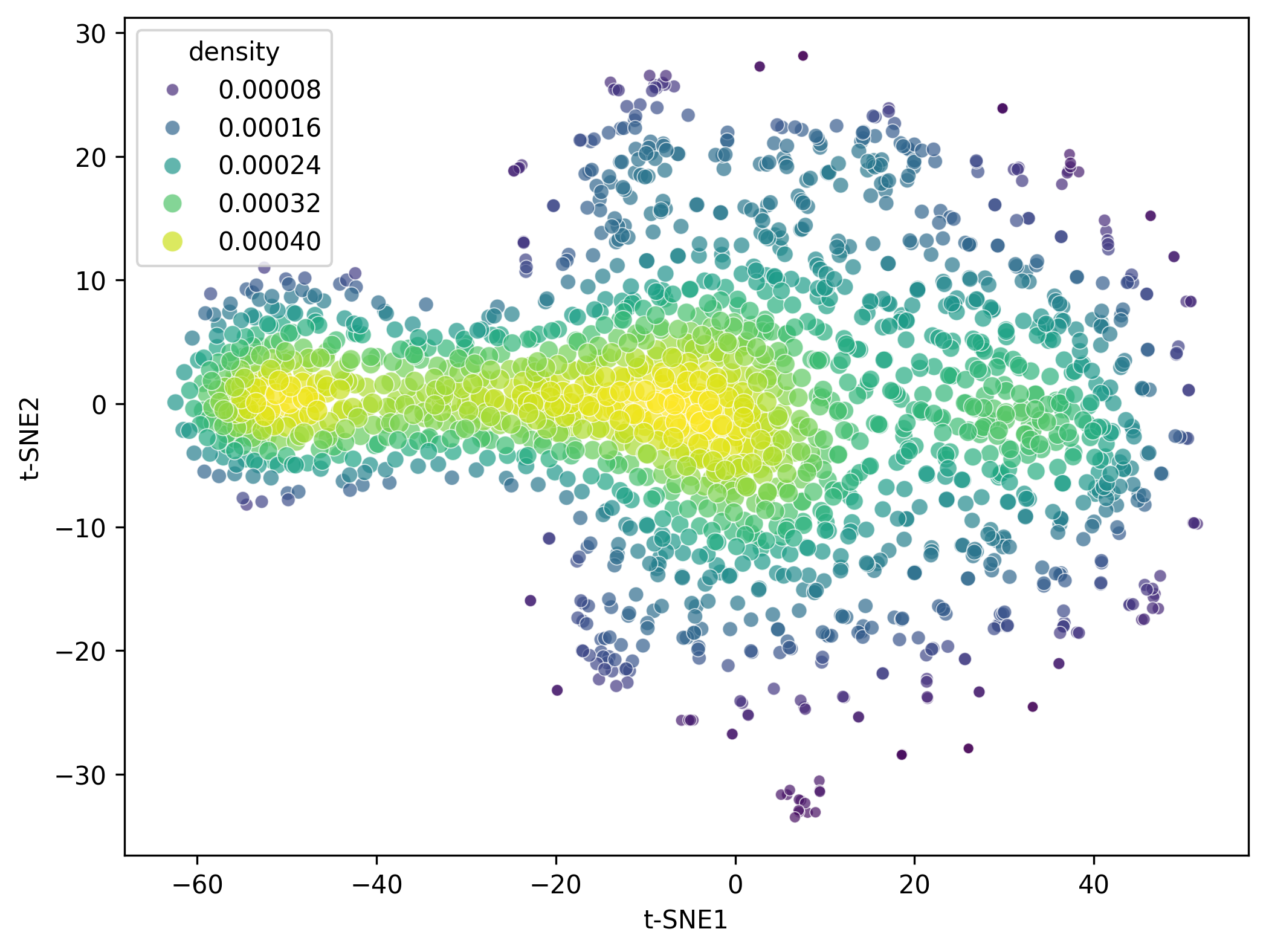}\label{fig:method_density}}
    \caption{Density visualisations of embedding spaces. \ETHOS~shows a more uniform spherical distribution (a), while \METHOD~develops a manifold structure with varying density regions (b), indicating better differentiation between disease categories.}
    \label{fig:density_embeddings}
\end{figure}

\subsubsection{Comparative Analysis of Embedding Structures}

To evaluate \METHOD's capacity for preserving clinical relationships, we conducted a comprehensive embedding analysis through multiple visualisation and clustering techniques. Figure \ref{fig:tsne_embeddings} presents the t-SNE projections of ICD code embeddings, revealing distinct organisational patterns between \ETHOS~and \METHOD.

\subsubsection{Global Structure Analysis}
\begin{figure}[t]
    \centering
    \subfloat[DBSCAN Clustering on ETHOS]{\includegraphics[width=0.48\columnwidth]{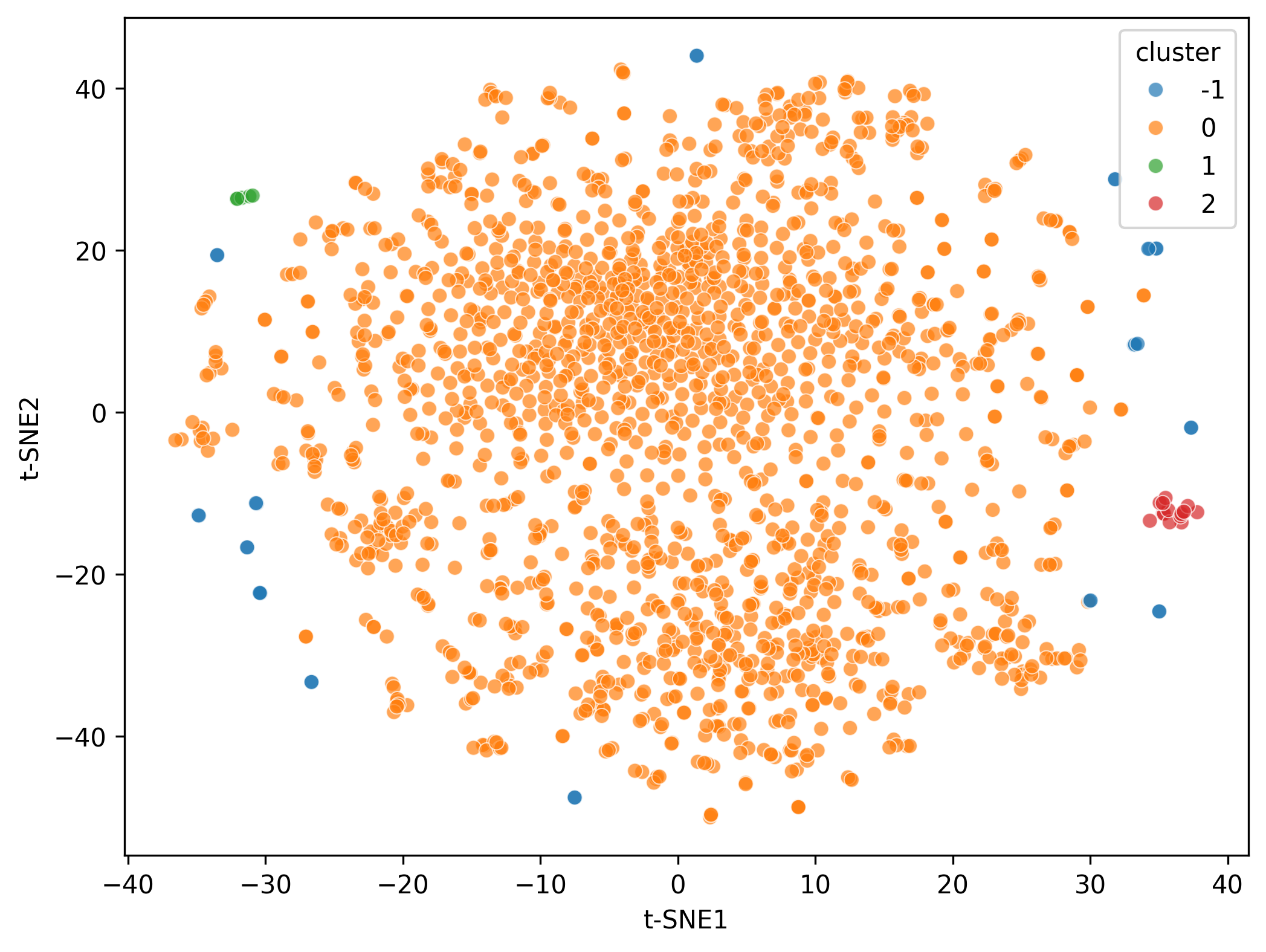}\label{fig:ethos_DBSCAN}}
    \hfill
    \subfloat[DBSCAN Clustering on METHOD]{\includegraphics[width=0.48\columnwidth]{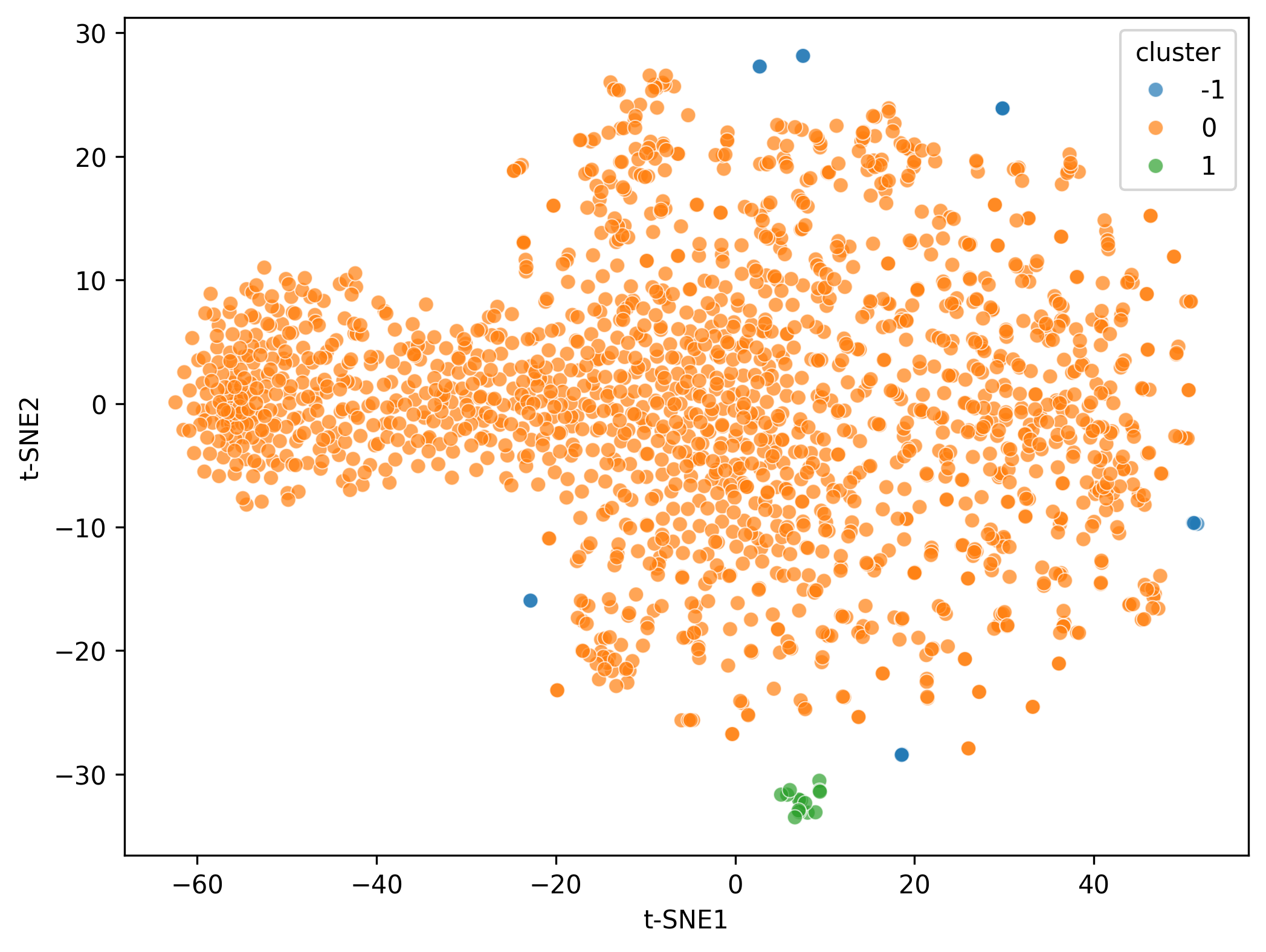}\label{fig:method_DBSCAN}}
    \caption{DBSCAN clustering results reveal more distinct and clinically meaningful clusters in \METHOD's embedding space (b) compared to \ETHOS~(a), with clusters corresponding more closely to clinical disease categories.}
    \label{fig:DBSCAN_embeddings}
\end{figure}

The density visualisations in Figure \ref{fig:density_embeddings} illustrate fundamental differences in how the two models structure medical knowledge:
\begin{itemize}
\item \textbf{\ETHOS} exhibits a spherical distribution with uniform density gradients, suggesting a generalised approach to medical concept representation. While this ensures broad coverage of clinical relationships, it may oversimplify the complex hierarchical nature of medical knowledge.
\item \textbf{\METHOD} develops a manifold structure with varying density regions, indicating better preservation of clinical hierarchies. The emergence of distinct high-density clusters (shown in Figure \ref{fig:DBSCAN_embeddings}) suggests the model has learnt to differentiate between major disease categories while maintaining relevant cross-category relationships.
\end{itemize}

\subsubsection{Local Pattern Analysis}
\begin{figure}[t]
    \centering
    \subfloat[High-Density ICD Code (ETHOS)]{\includegraphics[width=0.48\columnwidth]{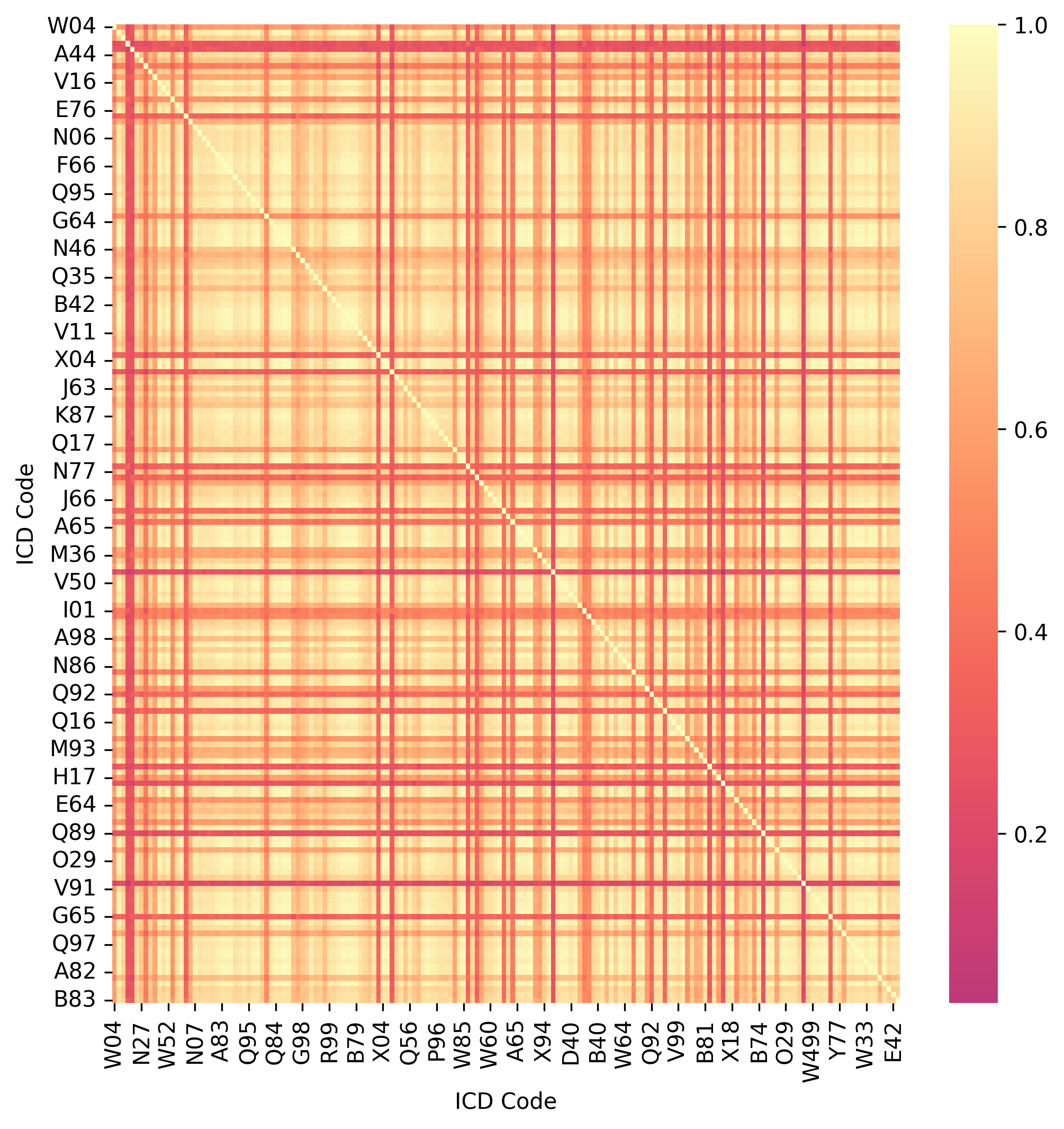}\label{fig:high_density_similarity_ethos}}
    \hfill
    \subfloat[High-Density ICD Code (METHOD)]{\includegraphics[width=0.48\columnwidth]{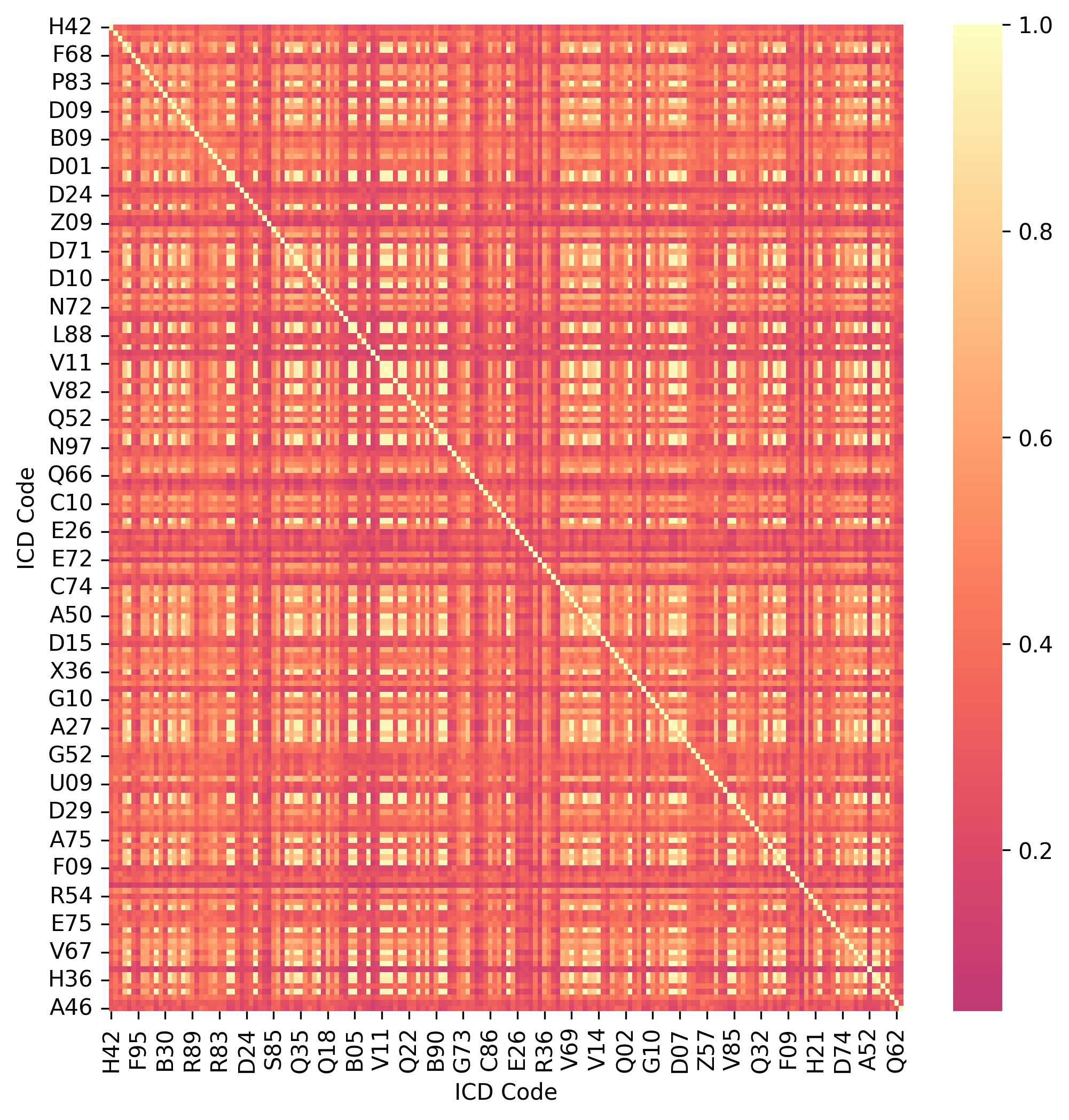}\label{fig:high_density_similarity_method}}
    \caption{Similarity heatmaps for high-density ICD code regions in \ETHOS~(a) and \METHOD~(b). \METHOD~demonstrates stronger intra-category similarities, particularly for related conditions within the same disease category.}
    \label{fig:heatmaps}
\end{figure}

\begin{figure}[t]
    \centering
    \includegraphics[width=0.6\columnwidth]{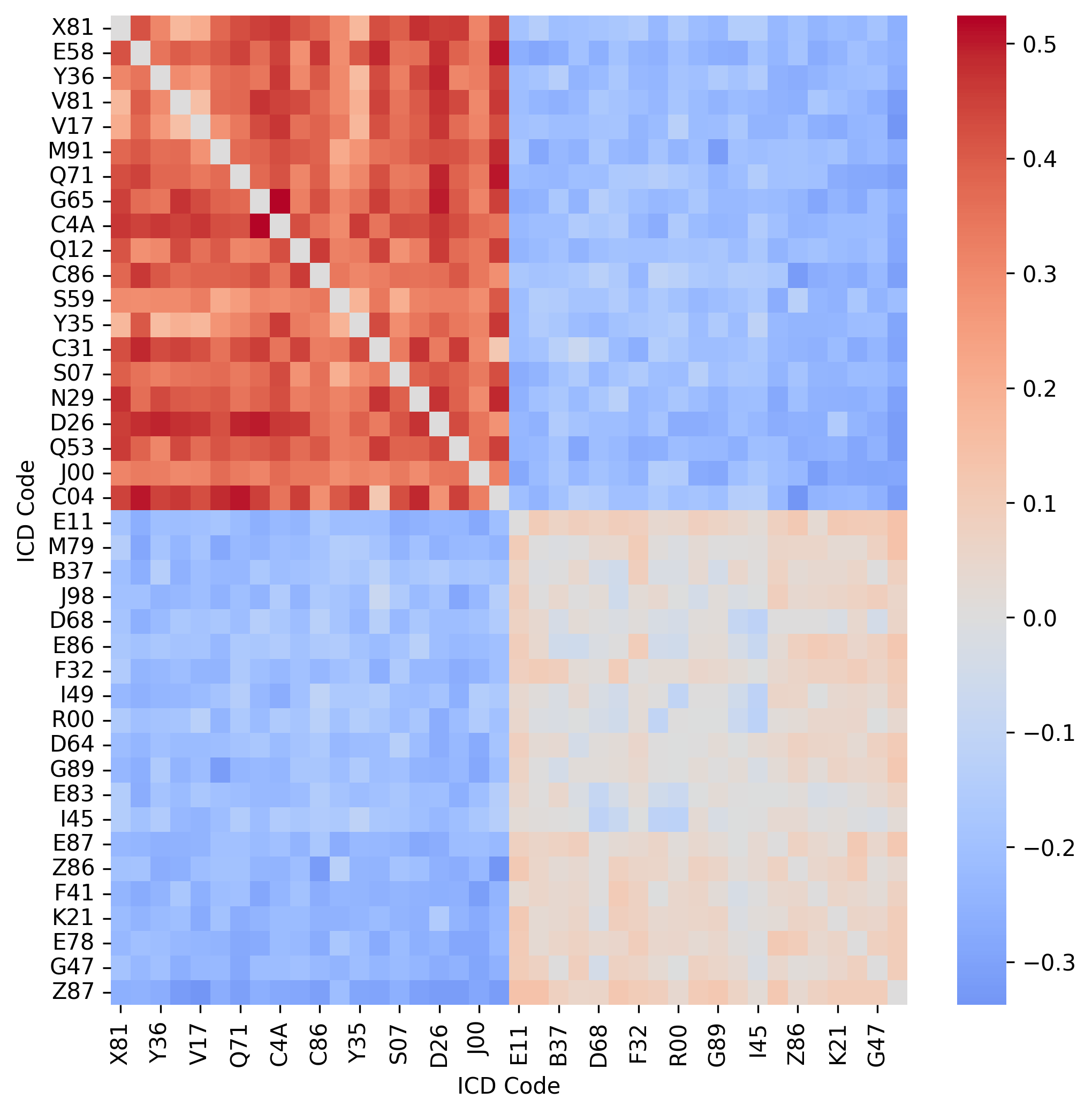}
    \caption{ICD Code Similarity Difference (\METHOD~- \ETHOS). Red indicates relationships that are stronger in \METHOD, while blue indicates relationships that are stronger in \ETHOS. The structured pattern suggests \METHOD~selectively enhances clinically relevant relationships.}
    \label{fig:similarity_diff_heatmap}
\end{figure}

The similarity heatmaps in Figure \ref{fig:heatmaps} reveal fine-grained differences in how the models encode clinical relationships:
\begin{itemize}
\item Within high-density regions, \METHOD~demonstrates stronger intra-category similarities, particularly for closely related conditions within the same ICD chapter.
\item The similarity difference heatmap (Figure \ref{fig:similarity_diff_heatmap}) shows \METHOD~selectively strengthens certain clinical associations while weakening others, potentially reflecting real-world medical knowledge structure.
\end{itemize}

\subsubsection{Clinical Implications}
These structural differences suggest both potential advantages and risks:
\begin{itemize}
\item \METHOD's more structured embedding space may improve accuracy in tasks requiring fine-grained clinical discrimination, such as specific disease prediction or comorbidity analysis.
\item However, the stronger clustering patterns could potentially lead to over-segmentation of the clinical space, making it important to validate the model's generalisation capabilities across different medical contexts.
\item The observed patterns warrant further investigation into whether the enhanced structural organisation genuinely reflects meaningful clinical relationships or introduces unwanted biases in medical concept representation.
\end{itemize}

This analysis provides insights into how architectural choices influence medical knowledge representation, though the clinical significance of these differences requires validation through downstream task performance and expert evaluation.

\section{Conclusion}

This paper has introduced \METHOD, a specialised transformer-based framework for clinical sequence modelling that addresses key challenges in healthcare AI. By optimising for long-term dependencies, mitigating data heterogeneity, and enhancing computational efficiency, \METHOD~demonstrates significant improvements over conventional architectures.

\subsection{Summary of Contributions}

Our research makes several significant contributions to clinical sequence modelling:

\begin{itemize}
    \item We introduce a patient-aware attention mechanism that ensures strict isolation of patient information while enabling efficient batch processing, providing theoretical guarantees for patient privacy and context preservation.
    
    \item We develop an adaptive sliding window attention scheme that effectively captures multi-scale temporal dependencies in irregularly sampled clinical data, reducing computational complexity from $O(n^2)$ to $O(nw)$.
    
    \item We design a U-Net inspired architecture with dynamic skip connections that better preserves both short-term and long-term clinical patterns, showing particular advantages for high-severity cases.
    
    \item We provide a comprehensive evaluation framework that assesses both computational performance and clinical relevance, revealing important insights about the relationship between model architecture and prediction reliability.
    
    \item We demonstrate empirically that \METHOD~outperforms existing approaches, particularly for high-severity cases that require urgent clinical intervention.
\end{itemize}

\subsection{Clinical Implications}

The experimental results highlight \METHOD's ability to balance predictive accuracy and clinical relevance. The patient-aware attention mechanism prevents cross-patient information leakage, while sliding window attention and multi-scale processing improve temporal representation, making it more effective for handling irregularly sampled medical data. These innovations align with broader efforts in medical AI to develop models that accommodate the complexity of patient trajectories without sacrificing interpretability.

The superior performance on high-severity cases (SOFA > 7) is particularly notable from a clinical perspective, as these represent the patients who most urgently require accurate prognostication and intervention. The ability to maintain consistent performance across varying sequence lengths addresses a critical challenge in clinical deployment, where patient histories vary significantly in length and granularity.

\subsection{Limitations and Future Work}

Despite these advancements, several challenges remain. The observed misalignment between token-level and continuous performance metrics suggests fundamental limitations in the current tokenisation approach. Future work should explore alternative discretisation strategies that better preserve clinically significant thresholds and physiological relationships.

The computational demands of deeper architectures also pose deployment concerns in resource-limited clinical environments. While the 12-layer model demonstrates superior performance, the trade-off between prediction accuracy and computational efficiency warrants careful consideration in practical applications.

Additional areas for future research include:

\begin{itemize}
    \item Exploring variable-specific tokenisation strategies that adapt to the unique distributions and clinical significance thresholds of different medical measurements
    
    \item Developing hybrid approaches that combine discrete tokens with continuous embeddings to better preserve fine-grained physiological relationships
    
    \item Extending the model to support multi-modal clinical data, including structured measurements, medical imaging, and unstructured clinical notes
    
    \item Conducting prospective clinical validation studies to assess the model's impact on real-world clinical decision-making and patient outcomes
    
    \item Enhancing interpretability through attention visualisation techniques that provide clinicians with transparent insights into the model's predictions
\end{itemize}

\subsection{Broader Impact}

\METHOD~represents a significant step towards transformer architectures optimised for healthcare applications, offering a foundation for future research into clinically grounded AI models. By addressing the unique challenges of medical sequence modelling—from irregular sampling to patient privacy concerns—this work contributes to the development of more reliable and clinically relevant predictive models.

As healthcare systems increasingly incorporate AI-driven decision support, architectures like \METHOD~that maintain both computational efficiency and clinical relevance will play a crucial role in translating the potential of modern transformer innovations into meaningful improvements in patient care.

\bibliographystyle{IEEEtran}

\begin{thebibliography}{10}

\bibitem{renc2024zero}
M. Renc, M. Hoffman, A. Shen, D. Hadfield-Menell, and Z. Lipton, ``Zero-shot multi-horizon clinical outcome prediction with large language models,'' in \textit{Proceedings of the 41st International Conference on Machine Learning}, 2024, pp. 29071--29087.

\bibitem{dao2022flashattention}
T. Dao, D. Fu, S. Ermon, A. Rudra, and C. Ré, ``FlashAttention: Fast and memory-efficient exact attention with IO-awareness,'' in \textit{Advances in Neural Information Processing Systems}, 2022, vol. 35, pp. 16344--16359.

\bibitem{Reproducibility}
M. McDermott, S. Wang, N. Marinsek, R. Ranganath, L. Foschini, and M. Ghassemi, ``Reproducibility in machine learning for health research: Still a ways to go,'' in \textit{Science Translational Medicine}, 2021, vol. 13, no. 586, pp. eabb1655.

\end{thebibliography}

\end{document}